\documentclass{article}
\usepackage[table]{xcolor}

\usepackage[main,final]{neurips_2026}

\usepackage{array}
\usepackage{tabularx}
\usepackage{placeins}

\newcommand{\swiftgroup}[2]{%
  \rowcolor{SwiftGroup}%
  \multicolumn{#1}{@{}l}{%
    \rule{0pt}{11.5pt}\hspace{3pt}\textit{#2}%
  }\\[-1pt]
}
\usepackage[utf8]{inputenc} 
\usepackage[T1]{fontenc}    
\usepackage{hyperref}       
\usepackage{url}            
\usepackage{booktabs}       
\usepackage{amsfonts}       
\usepackage{nicefrac}       
\usepackage{microtype}      
\usepackage{graphicx}
\usepackage{subcaption}
\usepackage{amsmath}
\usepackage{amssymb}
\usepackage{mathtools}
\usepackage{amsthm}
\usepackage{wrapfig}
\newtheorem{theorem}{Theorem}[section]
\theoremstyle{definition}
\newtheorem{definition}[theorem]{Definition}
\usepackage[capitalize,noabbrev]{cleveref}
\title{Breaking Information Islands in Sparse Tuning via Small-World Connectivity}

\author{%
  \textbf{Xianchao Guan}$^{1,2}$ \quad
  \textbf{Zijun Xiong}$^{1}$ \quad
  \textbf{Yifeng Wang}$^{3}$ \quad
  \textbf{Yubo Cui}$^{1}$ \\
  \textbf{Yaowei Wang}$^{1,2}$ \quad
  \textbf{Guo-Sen Xie}$^{4}$ \quad
  \textbf{Xin Li}$^{2}$%
  \thanks{Corresponding authors: Xin Li(xinlihitsz@gmail.com) and Zheng Zhang(darrenzz219@gmail.com).} \quad
  \textbf{Zheng Zhang}$^{1,2}$\footnotemark[1] \\[0.5em]
  $^{1}$Harbin Institute of Technology, Shenzhen, China \\
  $^{2}$Pengcheng Laboratory, China \\
  $^{3}$Tsinghua University, China \\
  $^{4}$Nanjing University of Science and Technology, China
}
\hypersetup{
  hidelinks,
  pdftitle={Breaking Information Islands in Sparse Tuning via Small-World Connectivity},
  pdfauthor={Xianchao Guan, Zijun Xiong, Yifeng Wang, Yubo Cui, Yaowei Wang, Guo-Sen Xie, Xin Li, Zheng Zhang},
  pdfsubject={Parameter-efficient fine-tuning with small-world connectivity},
  pdfkeywords={parameter-efficient fine-tuning, sparse adaptation, small-world connectivity, SWIFT}
}
\definecolor{SwiftRow}{rgb}{0.886275,0.925490,0.976471}
\definecolor{SwiftGroup}{rgb}{0.945098,0.913725,0.905882}
\definecolor{SwiftMuted}{rgb}{0.349020,0.388235,0.431373}

\begin{document}

\maketitle

\begin{abstract}
Sparse tuning is widely used to adapt large language models due to its parameter efficiency. However, its effectiveness depends not only on the number of trainable parameters, but also on the feature-interaction topology of the sparse update's structural support. Limited or uneven connectivity in this support may isolate subsets of feature dimensions, forming information islands that restrict direct task-specific interactions through the trainable update. To address this issue, we propose \textbf{S}mall-\textbf{W}orld \textbf{I}nduced \textbf{F}ine-\textbf{T}uning (\textbf{SWIFT}), a sparse tuning framework inspired by small-world connectivity. SWIFT combines block-diagonal sparse updates for local feature mixing with a fixed deterministic scattering permutation that routes the mixed outputs across feature groups without additional trainable parameters. We further show theoretically that certain random sparse-update models and block-diagonal updates can exhibit information islands, whereas a sufficiently scattering permutation guarantees positive block-level expansion when the trainable block-diagonal factor consists of equal-sized, structurally dense square blocks. Extensive experiments on commonsense reasoning, natural language generation, and image classification demonstrate that SWIFT outperforms the evaluated parameter-efficient fine-tuning (PEFT) baselines on average and can match or surpass full fine-tuning. Implementation is available at: \url{https://github.com/xianchaoguan/SWIFT}.
\end{abstract}

\section{Introduction}
Large language models (LLMs) have shown strong capabilities across language understanding and generation tasks \citep{kuang2025natural,axelsson2023using,DBLP:journals/pami/MengLHLWZS26,DBLP:journals/corr/abs-2506-21137}, accelerating the practical deployment of modern AI systems \citep{yang2024qwen2,jiang2026rethinking}. A common adaptation paradigm is to pre-train on large corpora and then fine-tune the model for specific tasks or domains \citep{zhang2023instruction,parthasarathy2024ultimate}. However, as model scale increases, full fine-tuning becomes costly in computation, memory, and deployment, limiting its scalability in practice \citep{zhao2024galore}. This has motivated parameter-efficient fine-tuning (PEFT), which aims to retain the adaptation ability of full fine-tuning while substantially reducing training and deployment costs \citep{kaplan2020scaling,zhao2024loraretriever,yang2024corda}.

\begin{figure}[htbp]
\centering
\includegraphics[width=\columnwidth]{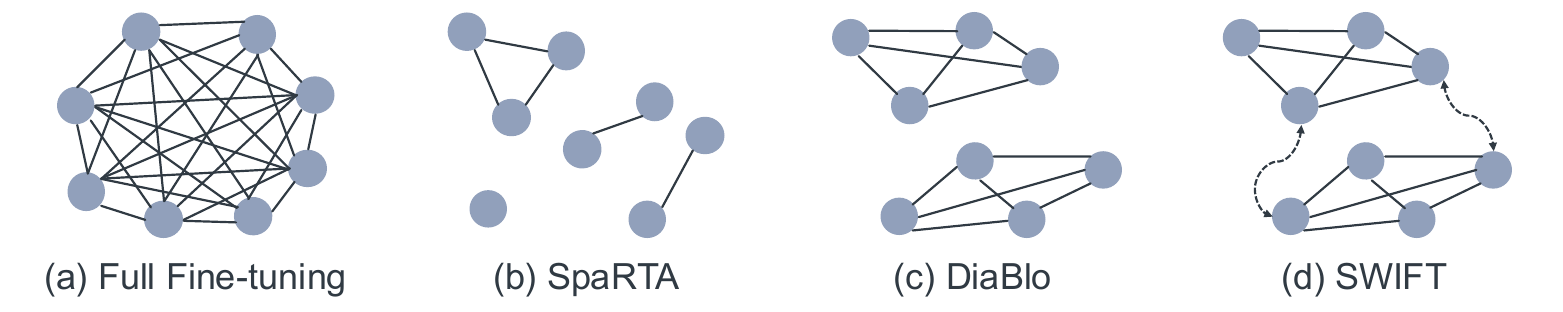}
\caption{Schematic feature-interaction topologies of trainable updates in full fine-tuning, SpaRTA, DiaBlo, and SWIFT. SWIFT combines local feature mixing with permutation-based routing across feature groups.}
\label{Figure1}
\end{figure}

PEFT methods adapt large models by updating only a small set of parameters \citep{zhang2025parameter}. Low-Rank Adaptation (LoRA) \citep{hu2022lora} is a representative approach, parameterizing weight updates with low-rank matrices while freezing pre-trained weights. Its variants further improve expressiveness, initialization, and optimization behavior \citep{zhang2023adalora,liu2024dora,meng2024pissa,jiang2025dmlora}. However, low-rank methods are inherently constrained by rank bottlenecks, which may limit expressive capacity and downstream performance \citep{hayou2024lora+,he2025rasa}. Sparse tuning offers an alternative by directly updating a discrete subset of pre-trained weights, maintaining parameter efficiency without imposing an explicit low-rank factorization \citep{song2023sparse,yang2024s,wang2025parameter}. Recent structured update parameterizations further suggest that the update structure itself is crucial for efficient adaptation \citep{tan2024more,gurses2025diablo}. Yet how such structures shape information propagation across feature dimensions remains insufficiently understood.

From the perspective of forward computation, neural network inference can be viewed as information propagation in a high-dimensional feature space \citep{joshi2025transformers}. Full fine-tuning allows dense weight updates, enabling direct task-specific interactions across feature dimensions, as illustrated in Figure~\ref{Figure1}(a). Sparse tuning, in contrast, updates only a subset of entries or structured components, inducing a sparse feature-interaction graph defined by the update's structurally trainable support. Limited or uneven connectivity in this graph may leave subsets of feature dimensions weakly connected or isolated, forming information islands that restrict direct task-specific interactions through the trainable update. Such islands arise in purely block-diagonal methods such as DiaBlo \citep{gurses2025diablo}, which preserve local interactions but lack trainable cross-block communication (Figure~\ref{Figure1}(c)). Irregular or random sparse methods such as SpaRTA \citep{aliaga2025sparsity} may also exhibit uneven feature-group coverage depending on the sampled support (Figure~\ref{Figure1}(b)).

This observation suggests that the effectiveness of sparse tuning depends not only on the number of trainable parameters, but also on how well the update supports global communication across feature dimensions. Small-world connectivity provides a natural principle for this goal, combining strong local interactions with sparse long-range shortcuts to enable efficient information exchange without dense connectivity \citep{lachgar2024uncovering,morais2025simple}. In this context, an effective sparse update should combine local feature mixing with reliable access to distant feature groups, motivating designs that integrate structured local mixing with distributed long-range routing.

In this paper, we propose \textbf{S}mall-\textbf{W}orld \textbf{I}nduced \textbf{F}ine-\textbf{T}uning (\textbf{SWIFT}), a sparse tuning framework inspired by small-world connectivity (Figure~\ref{Figure1}(d)). SWIFT uses block-diagonal sparse updates for local feature mixing and applies a fixed deterministic scattering permutation to route the mixed outputs across feature groups without additional trainable parameters. Theoretically, we show that certain random sparse-update models and block-diagonal updates can exhibit information islands, whereas a sufficiently scattering permutation guarantees positive block-level expansion when the trainable block-diagonal factor consists of equal-sized, structurally dense square blocks. Extensive experiments on commonsense reasoning, natural language generation, and image classification demonstrate that SWIFT outperforms the evaluated PEFT baselines on average and can match or surpass full fine-tuning.

\section{Related Work}
\subsection{Low-Rank Adaptation}
Low-rank adaptation is a widely used PEFT paradigm due to its simplicity and strong empirical performance. LoRA \citep{hu2022lora} enables efficient adaptation by injecting trainable low-rank matrices while freezing pre-trained weights. Subsequent work improves its expressiveness and optimization through better parameterization, initialization, and architectural design, including DoRA \citep{liu2024dora}, PiSSA \citep{meng2024pissa}, HydraLoRA \citep{tian2024hydralora}, and GOAT \citep{fan2025make}. However, low-rank factorization restricts the rank of the parameterized update, which may limit the ability to model complex cross-dimensional interactions as adaptation demands increase \citep{zeng2023expressive,hayou2024lora+}.

\subsection{Sparse Fine-Tuning}
Sparse fine-tuning directly updates a subset of pre-trained weights, maintaining parameter efficiency without imposing an explicit low-rank factorization \citep{song2023sparse,yang2024s,wang2025parameter}. Existing methods differ in how the sparse update is constructed. Data-driven approaches select task-relevant parameters based on importance signals, such as SMT \citep{he2025smt} and SpIEL \citep{ansell2024scaling}. In contrast, SpaRTA \citep{aliaga2025sparsity} randomly selects a fixed subset of parameters for adaptation. S$^{2}$FT \citep{yang2024s} selects attention heads and FFN channels, then co-permutes coupled weight matrices to organize the selected components into dense submatrices for efficient updates. Related structured parameterizations further highlight the importance of update design: MoRe \citep{tan2024more} uses two trainable block-diagonal factors and two fixed permutations in a Monarch-style parameterization, while DiaBlo \citep{gurses2025diablo} directly updates dense diagonal blocks.

SWIFT combines a single trainable block-diagonal factor with a fixed deterministic scattering permutation, yielding $\Delta W = BP$. This composition routes locally mixed outputs across feature groups without additional trainable parameters. Alongside this update design, we analyze how an explicit scattering condition controls remote-group coverage and expansion under stated structural assumptions.

\section{Method}
We present SWIFT, which combines block-diagonal updates with a fixed deterministic scattering permutation to route locally mixed outputs across feature groups. We then introduce Minimum Remote Group Coverage (MRGC), a method-agnostic measure of connectivity in the structural support of trainable updates, focusing on worst-case access to remote groups. Finally, we provide a graph-theoretic analysis of information islands and establish coverage and expansion guarantees under explicit structural and scattering assumptions.

\subsection{SWIFT Framework}
\begin{figure}[htbp]
\centering
\includegraphics[width=\textwidth]{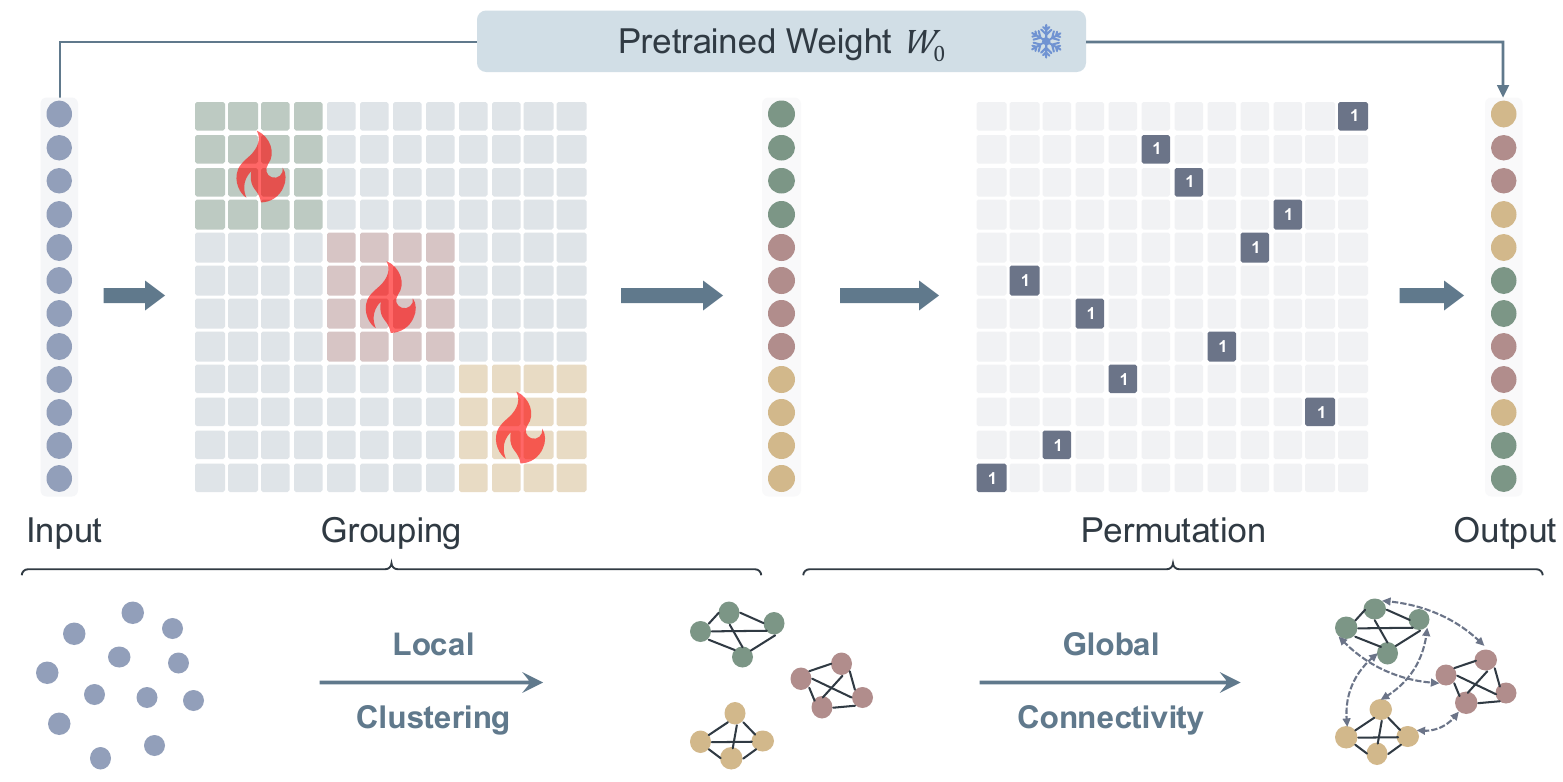}
\caption{Overview of the SWIFT framework. SWIFT combines trainable block-diagonal updates for local feature mixing with a fixed deterministic scattering permutation that routes the mixed outputs across feature groups. The pre-trained weights remain frozen.}
\label{Figure2}
\end{figure}

SWIFT is applied to standard linear modules in Transformers, such as query/key/value projections and feed-forward network layers. Consider a frozen pre-trained linear weight $\mathbf{W}\in\mathbb{R}^{m_1\times m_2}$ in such a module. Given an input activation $\mathbf{x}\in\mathbb{R}^{1\times m_1}$, the original computation is
\begin{equation}
\mathbf{y}=\mathbf{x}\mathbf{W}.
\label{eq:base_linear}
\end{equation}
SWIFT introduces a sparse trainable update that combines local feature mixing with routing across feature groups. Specifically, we parameterize the effective update as a block-diagonal trainable matrix followed by a fixed permutation:
\begin{equation}
\mathbf{y}
=
\mathbf{x}\mathbf{W}
+
\mathbf{x}\mathbf{B}\mathbf{P},
\label{eq:swift}
\end{equation}
where $\mathbf{B}\in\mathbb{R}^{m_1\times m_2}$ is trainable and block diagonal, and $\mathbf{P}\in\mathbb{R}^{m_2\times m_2}$ is fixed throughout fine-tuning. Within each adapted linear module, only $\mathbf{B}$ is trainable. We initialize $\mathbf{B}$ to zero, preserving the original module's output at initialization. The block structure of $\mathbf{B}$ provides local feature mixing, while $\mathbf{P}$ routes the mixed outputs across feature groups. Thus, the effective update $\mathbf{B}\mathbf{P}$ introduces cross-group interactions without increasing the number of trainable parameters, as illustrated in Figure~\ref{Figure2}.

\paragraph{Grouping: block-diagonal sparse updates.}
SWIFT first partitions the input and output feature dimensions of a linear module into $N>1$ groups. When $N$ divides both $m_1$ and $m_2$, let $d_1=m_1/N$ and $d_2=m_2/N$. We divide $\mathbf{B}$ into $N\times N$ blocks $\mathbf{B}_{ij}\in\mathbb{R}^{d_1\times d_2}$ and enforce the block-diagonal structure
\begin{equation}
\mathbf{B}=
\begin{pmatrix}
\mathbf{B}_{11} & \mathbf{0}      & \cdots & \mathbf{0}\\
\mathbf{0}      & \mathbf{B}_{22} & \cdots & \mathbf{0}\\
\vdots         & \vdots         & \ddots & \vdots\\
\mathbf{0}      & \mathbf{0}      & \cdots & \mathbf{B}_{NN}
\end{pmatrix}.
\label{eq:blockdiag}
\end{equation}
During fine-tuning, all entries within the diagonal blocks $\{\mathbf{B}_{ii}\}_{i=1}^{N}$ are trainable, while the off-diagonal blocks remain zero. The number of trainable parameters is $Nd_1d_2=m_1m_2/N$. This structure concentrates the trainable budget on dense within-group interactions, providing efficient local adaptation. By itself, a block-diagonal update has no trainable interactions between differently indexed input and output groups.

\paragraph{Permutation: long-range routing.}
To enable cross-group interactions without introducing additional trainable parameters, SWIFT composes the block-diagonal update with a fixed permutation $\mathbf{P}$. Right multiplication by $\mathbf{P}$ reorders the output coordinates of $\mathbf{x}\mathbf{B}$, dispersing each block's mixed outputs across feature groups. Consequently, the effective update $\mathbf{B}\mathbf{P}$ connects input groups to dispersed output coordinates under the same parameter budget. The permutation is determined by the layer dimensions and block number, remains fixed throughout training, and requires no optimizer state.

\paragraph{Efficient implementation.}
In practice, $\mathbf{P}$ is implemented by index reordering and is never explicitly materialized as a dense matrix. We generate its index array using a reshape--transpose--reshape procedure: output indices are reshaped into an $N\times d_2$ grid, transposed, and flattened. The resulting array is used to gather output coordinates. Writing $\pi(j)$ for the pre-permutation coordinate selected for output coordinate $j$, our convention is
\begin{equation}
(\mathbf{B}\mathbf{P})_{ij}
=
\mathbf{B}_{i,\pi(j)}.
\label{eq:permutation_convention}
\end{equation}

When $N$ divides the output width, the number of coordinates routed from any one intermediate output group to any one final output group is either $\lfloor d_2/N\rfloor$ or $\lceil d_2/N\rceil$. Each group therefore reaches exactly $\min(N,d_2)$ target groups. Exact uniform scattering requires $N\mid d_2$, equivalently $N^2\mid m_2$. Otherwise, the pairwise loads differ by at most one coordinate, with realized scattering bound
\begin{equation}
\epsilon_{\mathrm{real}}
=
\frac{\lceil d_2/N\rceil}{d_2}.
\label{eq:real_scattering}
\end{equation}

We use $N=64$ in all main experiments, for which the input and output widths of all adapted projections are divisible by $N$. For incompatible dimensions, the implementation pads the input and block dimensions to the next multiples of $N$, applies the permutation at the padded output width, and retains the first $m_2$ output coordinates. In this case, coverage and scattering are evaluated on the retained support after slicing.

\subsection{Global Connectivity and Remote Group Coverage Analysis}

The connectivity induced by the effective update is key to understanding sparse fine-tuning. We analyze its \emph{structural support}: the set of positions at which the update is allowed to vary, including trainable entries initialized to zero. This support determines the possible direct task-specific interactions through the trainable update. A useful measure should therefore capture not only sparsity, but also whether each feature dimension has sufficient access to remote feature groups.

To this end, we introduce \emph{Minimum Remote Group Coverage} (MRGC), a feature-wise worst-case measure of remote-group connectivity. Consider a square linear layer adapted as $\mathbf{y}=\mathbf{x}(\mathbf{W}+\Delta\mathbf{W})$, where $\mathbf{W}$ is frozen and $\Delta\mathbf{W}$ is the effective update. Let $\Omega_{\Delta}\subseteq\{1,\ldots,d\}\times\{1,\ldots,d\}$ denote the structural support of $\Delta\mathbf{W}$. Each pair $(i,j)\in\Omega_{\Delta}$ represents a possible direct trainable interaction from input feature dimension $i$ to output feature dimension $j$.

Let $V=\{1,\ldots,d\}$ be partitioned into $K>1$ groups, $\mathcal{P}=\{V_1,\ldots,V_K\}$, and let $g(i)$ denote the index of the group containing feature dimension $i$. For each source feature dimension $i$, we define the set of directly reachable remote target groups as
\begin{equation}
\mathcal{R}(i)
=
\left\{
q\in\{1,\ldots,K\}\setminus\{g(i)\}:
\exists j\in V_q,\ (i,j)\in\Omega_{\Delta}
\right\}.
\label{eq:2}
\end{equation}
The MRGC score is defined as
\begin{equation}
\mathrm{MRGC}
=
\min_{i\in V}
\frac{|\mathcal{R}(i)|}{K-1}.
\label{eq:3}
\end{equation}

MRGC measures the worst-case fraction of remote feature groups that each feature dimension can directly reach through the update. Poorly covered features may become bottlenecks for direct cross-group interactions. By excluding within-group interactions, MRGC does not reward purely local updates and, under a shared partition and matched density, provides a method-agnostic measure of remote-group coverage.

At $d=4096$ and $K=64$, SWIFT attains an MRGC of $1.00$ at density $1/64$, compared with $0.00$ for block-diagonal tuning and a mean of $0.41$ over $1{,}000$ random masks at the same expected density; dense full fine-tuning also attains $1.00$. Appendix~\ref{app:support_simulation} provides the simulation setup and the corresponding group-coverage definition for rectangular layers.

\subsection{Theory of Small-World Induced Fine-Tuning}
\label{3.2}

We provide a graph-theoretic analysis of connectivity in the structural support of trainable updates. For square layers, we identify input and output coordinates with a common vertex set. Under this view, information islands are subsets of feature dimensions with no outgoing trainable interactions to their complement. We analyze how local mixing and permutation-based routing affect this structural connectivity.

\begin{definition}[\textbf{Operator-Induced Interaction Graph}]
Let $\Delta\mathbf{W}\in\mathbb{R}^{d\times d}$ be a trainable update with structural support $\Omega_{\Delta}$. Its \emph{operator-induced interaction graph} is the directed graph $G_{\Delta\mathbf{W}}=(V,E)$ with $V=\{1,\ldots,d\}$ and
\begin{equation}
E=\Omega_{\Delta}.
\label{eq:op_graph}
\end{equation}
Under the row-vector convention $\mathbf{y}=\mathbf{x}(\mathbf{W}+\Delta\mathbf{W})$, an edge $(i,j)\in E$ represents a possible direct trainable interaction from input coordinate $i$ to output coordinate $j$.
\end{definition}

To quantify bottlenecks in this support graph, we use directed out-vertex expansion.

\begin{definition}[\textbf{Directed Out-Vertex Expansion}]
For a directed graph $G=(V,E)$ and a subset $S\subseteq V$, define
\[
O(S)=\{j\in V:\exists i\in S,\ (i,j)\in E\},
\qquad
\partial_{\mathrm{out}}S=O(S)\setminus S.
\]
The coordinate-level out-vertex expansion is
\begin{equation}
h(G)
\triangleq
\min_{\substack{S\subseteq V\\1\le |S|\le\lfloor d/2\rfloor}}
\frac{|\partial_{\mathrm{out}}S|}{|S|}.
\label{eq:cheeger}
\end{equation}
When $h(G)=0$, there exists a nonempty subset of at most half the coordinates with no outgoing edges to its complement in the update-support graph.
\end{definition}

\begin{theorem}[\textbf{Connectivity Collapse under Independent Random Sparsification}]
\label{thm:nonregular}
Fix a constant $a>0$. Consider a sequence of random structural supports on $d$ vertices whose outgoing supports are sampled independently across vertices and whose expected average out-degree tends to $a$. Assume that
\[
\liminf_{d\to\infty}
\min_{v\in\{1,\ldots,d\}}
\mathbb{P}\bigl(D_{\mathrm{out}}(v)=0\bigr)
>0.
\]
Then the induced graph $G_d$ contains zero-out-degree vertices with probability approaching $1$, and
\begin{equation}
\lim_{d\to\infty}
\mathbb{P}\bigl(h(G_d)=0\bigr)
=1.
\label{theorem1}
\end{equation}
\end{theorem}

Theorem~\ref{thm:nonregular} identifies a failure mode of random sparse updates in the constant-expected-degree regime. For example, independent Bernoulli sampling with entry probability $a/d$ satisfies the assumption because the zero-out-degree probability is $(1-a/d)^d\to e^{-a}>0$. A zero-out-degree vertex forms a singleton with no outgoing boundary. The fixed-density MRGC simulation above examines uneven group coverage at a different sparsity scale. A detailed proof is provided in Appendix~\ref{B1}.

\begin{theorem}[\textbf{Zero Expansion of Block-Diagonal Updates}]
\label{thm:block}
Let $\mathbf{B}\in\mathbb{R}^{d\times d}$ have block-diagonal structural support with $k>1$ equal-sized blocks. Then its induced graph satisfies
\begin{equation}
h(G_{\mathbf{B}})=0.
\end{equation}
\end{theorem}

Theorem~\ref{thm:block} formalizes the connectivity bottleneck of purely block-diagonal updates. Each block has no outgoing edges to other blocks, and its size is at most $d/2$. Thus, a block forms a subset with zero outgoing boundary. A detailed proof is provided in Appendix~\ref{B2}.

SWIFT addresses block isolation by composing $\mathbf{B}$ with a fixed permutation $\mathbf{P}$, which disperses locally mixed outputs across feature blocks. We capture this effect through a scattering condition.

\begin{definition}[\textbf{$\epsilon$-Uniform Scattering Permutation}]
\label{def:scatter}
Partition $V=\{1,\ldots,d\}$ into $k>1$ equal-sized blocks $\mathcal{P}=\{V_r\}_{r=1}^{k}$ with block size $m=d/k$. A permutation $\pi:V\to V$ is \emph{$\epsilon$-uniformly scattering} with respect to this partition if
\begin{equation}
|\pi(V_r)\cap V_q|
\le \epsilon m
\quad\text{for all }r,q\in\{1,\ldots,k\},
\qquad
\epsilon\in[1/k,1].
\label{eq:eps_scatter}
\end{equation}
We use the permutation-matrix convention $\mathbf{P}_{\pi(j),j}=1$, so $(\mathbf{B}\mathbf{P})_{ij}=\mathbf{B}_{i,\pi(j)}$. Smaller $\epsilon$ indicates more evenly dispersed block overlaps. Exact uniform scattering has $\epsilon=1/k$ and requires $k\mid m$.
\end{definition}

To characterize expansion across complete feature blocks, write $U_I=\bigcup_{r\in I}V_r$ and define the block-level out-vertex expansion as
\begin{equation}
h_{\mathcal{P}}(G)
\triangleq
\min_{\substack{I\subseteq\{1,\ldots,k\}\\
1\le |I|\le\lfloor k/2\rfloor}}
\frac{|O(U_I)\setminus U_I|}{|U_I|}.
\label{eq:block_expansion}
\end{equation}
Because block unions form a restricted family of coordinate subsets,
\[
h(G)\le h_{\mathcal{P}}(G).
\]
MRGC measures per-vertex remote-group coverage and does not account for neighborhood overlap across vertices, so it has no general ordering with either expansion measure.

\begin{theorem}[\textbf{Expansion of Small-World Induced Fine-Tuning}]
\label{thm:swift}
Let $\mathbf{B}\in\mathbb{R}^{d\times d}$ have $k>1$ equal-sized, structurally dense diagonal blocks of size $m=d/k$, with
\[
\Omega_{\mathbf{B}}
=
\bigcup_{r=1}^{k}(V_r\times V_r).
\]
Let $\mathbf{P}$ correspond to an $\epsilon$-uniformly scattering permutation with respect to the same partition. If $\epsilon<2/k$, then the structural support of the SWIFT update $\mathbf{B}\mathbf{P}$ satisfies
\begin{equation}
h_{\mathcal{P}}(G_{\mathbf{B}\mathbf{P}})
\ge
1-\frac{k\epsilon}{2}
>0.
\label{eq:swift_expansion}
\end{equation}
\end{theorem}

The same scattering condition separately yields MRGC and coordinate-level expansion bounds; their statements and proofs are provided in Appendix~\ref{app:additional_bounds} and Appendix~\ref{B3}. Under exact uniform scattering, $\epsilon=1/k$, the block-level lower bound is $1/2$.

Taken together, Theorems~\ref{thm:nonregular}--\ref{thm:swift} identify distinct connectivity properties of sparse update supports and support combining local feature mixing with distributed cross-group routing under a sparse parameter budget. The expansion guarantees apply under the stated assumptions for square layers; rectangular projections are characterized through bipartite group coverage.

\section{Experiments and Results}

\subsection{Baselines}

We compare SWIFT with full fine-tuning; low-rank methods LoRA \citep{hu2022lora}, DoRA \citep{liu2024dora}, PiSSA \citep{meng2024pissa}, HydraLoRA \citep{tian2024hydralora}, and GOAT \citep{fan2025make}; and sparse or structured methods SpaRTA \citep{aliaga2025sparsity}, S$^{2}$FT \citep{yang2024s}, SMT \citep{he2025smt}, SpIEL \citep{ansell2024scaling}, MoRe \citep{tan2024more}, and DiaBlo \citep{gurses2025diablo}. Full fine-tuning updates all pre-trained parameters. Baseline descriptions and task-specific settings are provided in Appendices~\ref{app:baselines} and~\ref{A2}.

\begin{table}[htbp]
\centering
\caption{Performance comparison of LLaMA3-8B on eight commonsense reasoning datasets. Task scores are accuracies (\%). The best and second-best scores in each column are indicated in \textbf{bold} and \underline{underlined}, respectively, including ties.}
\label{table1}
\begingroup
\fontsize{8}{11}\selectfont
\renewcommand{\arraystretch}{1.08}
\setlength{\tabcolsep}{1.4pt}
\begin{tabularx}{\linewidth}{@{}l*{10}{>{\centering\arraybackslash}X}@{}}
\toprule
\textbf{Method} & \textbf{\shortstack{Param.\\(\%)}} & \textbf{BoolQ} & \textbf{PIQA} & \textbf{SIQA} & \textbf{HellaS.} & \textbf{WinoG.} & \textbf{ARC-e} & \textbf{ARC-c} & \textbf{OBQA} & \textbf{Avg.} \\
\midrule
Full Fine-Tuning & 100 & 73.90 & 86.20 & 79.10 & 93.10 & 85.80 & 88.10 & 78.20 & 84.00 & 83.55 \\
\swiftgroup{11}{Low-rank adaptation}
LoRA & 0.70 & 74.58 & 88.41 & 80.81 & 95.31 & 85.87 & 89.60 & 80.12 & 85.40 & 85.01 \\
DoRA & 0.71 & \underline{76.09} & 88.30 & 80.60 & 95.23 & 86.67 & 90.40 & 80.29 & 88.20 & 85.72 \\
PiSSA & 0.70 & 69.70 & 87.20 & 79.50 & 90.30 & 81.00 & 82.90 & 75.50 & 77.80 & 80.49 \\
HydraLoRA & 0.53 & 72.20 & 85.00 & 79.80 & 89.60 & 87.60 & 79.40 & 74.00 & 80.40 & 81.00 \\
\swiftgroup{11}{Sparse and structured adaptation}
SpIEL & 0.70 & 72.10 & 83.60 & 80.00 & 91.80 & 85.40 & 91.20 & 76.80 & 80.80 & 82.71 \\
S$^{2}$FT & 0.70 & 75.00 & 89.00 & 80.70 & \textbf{96.50} & 88.00 & 92.50 & \underline{83.40} & 87.80 & 86.61 \\
MoRe & 0.70 & 71.93 & 89.28 & 82.24 & 95.22 & 88.79 & \underline{93.27} & 83.19 & 88.80 & 86.59 \\
SMT & 3.01 & 75.10 & \underline{89.90} & 82.40 & \underline{96.30} & 88.80 & 92.60 & 82.80 & \underline{89.60} & 87.19 \\
SpaRTA & 0.44 & 75.60 & \textbf{90.32} & \underline{82.91} & 95.06 & 89.03 & 93.18 & 82.94 & 89.20 & \underline{87.28} \\
DiaBlo & 0.88 & 74.70 & 89.20 & 82.70 & 95.20 & \underline{89.20} & 93.10 & 82.90 & 89.00 & 87.00 \\
\addlinespace[2pt]
\rowcolor{SwiftRow}
\textbf{SWIFT} & 0.88 & \textbf{76.30} & \textbf{90.32} & \textbf{83.31} & 95.90 & \textbf{89.27} & \textbf{93.52} & \textbf{84.18} & \textbf{91.60} & \textbf{88.05} \\
\bottomrule
\end{tabularx}
\endgroup
\end{table}

\subsection{Datasets}

We evaluate SWIFT across 17 datasets in three task categories. For commonsense reasoning, we fine-tune LLaMA3-8B \citep{grattafiori2024llama} on Commonsense-170K and evaluate on eight datasets \citep{hu2023llm}. For natural language generation, we separately fine-tune LLaMA2-7B \citep{touvron2023llama} on 100K examples each from MetaMathQA \citep{yu2023metamath} and CodeFeedback \citep{zheng2024opencodeinterpreter}, using the data prepared by PiSSA \citep{meng2024pissa}, and evaluate on GSM8K, MATH, HumanEval, and MBPP. For image classification, we fine-tune CLIP ViT-B/32 \citep{radford2021learning} on five datasets \citep{fan2025make}. Dataset descriptions, references, and experimental settings are provided in Appendices~\ref{A1} and~\ref{A2}.

\subsection{Comparative Results}

We compare SWIFT with full fine-tuning, LoRA variants, and sparse or structured fine-tuning methods across the three task categories. In Tables~\ref{table1}--\ref{table3}, Avg.\ denotes the arithmetic mean of the displayed task scores. Average-score differences reported below are calculated from the displayed averages.

\subsubsection{Commonsense Reasoning}

As shown in Table~\ref{table1}, SWIFT achieves the highest average accuracy of $88.05\%$ at a reported trainable-parameter ratio of $0.88\%$. It outperforms LoRA by 3.04 percentage points and the second-best method, SpaRTA, by 0.77 percentage points. SWIFT also surpasses DiaBlo by 1.05 percentage points, with both methods reporting a trainable-parameter ratio of $0.88\%$. This comparison supports the benefit of combining block-diagonal adaptation with permutation-based routing.

\subsubsection{Natural Language Generation}

As shown in Table~\ref{table2}, SWIFT achieves the highest average score of $32.56\%$ at a reported trainable-parameter ratio of $1.50\%$, surpassing full fine-tuning by 4.26 percentage points. It outperforms LoRA by 7.20 percentage points, the second-best method MoRe by 0.59 percentage points, and SpaRTA by 1.30 percentage points. SWIFT also exceeds DiaBlo by 1.03 percentage points, with both methods reporting a trainable-parameter ratio of $1.50\%$. SWIFT achieves the best results on GSM8K, MATH, and MBPP, demonstrating effective adaptation across mathematical reasoning and code generation.

\begin{table}[htbp]
\centering
\caption{Performance comparison of LLaMA2-7B on four natural language generation datasets. Scores are percentages. Bold and underlined entries indicate the best and second-best scores, respectively, including ties.}
\label{table2}
\begingroup
\fontsize{9}{11}\selectfont
\renewcommand{\arraystretch}{1.08}
\setlength{\tabcolsep}{3pt}
\begin{tabularx}{\linewidth}{@{}l*{6}{>{\centering\arraybackslash}X}@{}}
\toprule
\textbf{Method} & \textbf{\shortstack{Param.\\(\%)}} & \textbf{GSM8K} & \textbf{MATH} & \textbf{HumanEval} & \textbf{MBPP} & \textbf{Avg.} \\
\midrule
Full Fine-Tuning & 100 & 49.05 & 7.22 & 21.34 & 35.59 & 28.30 \\
\swiftgroup{7}{Low-rank adaptation}
LoRA & 4.75 & 42.30 & 5.50 & 18.29 & 35.34 & 25.36 \\
DoRA & 4.76 & 50.42 & 7.26 & 23.78 & \underline{38.85} & 30.08 \\
PiSSA & 4.75 & 53.07 & 7.44 & 21.95 & 37.09 & 29.89 \\
HydraLoRA & 1.87 & 49.58 & 7.02 & 23.17 & 37.34 & 29.28 \\
\swiftgroup{7}{Sparse and structured adaptation}
MoRe & 2.37 & \underline{53.82} & \underline{9.40} & \underline{26.83} & 37.84 & \underline{31.97} \\
SpaRTA & 0.76 & 53.50 & 8.00 & \textbf{27.44} & 36.09 & 31.26 \\
DiaBlo & 1.50 & 53.07 & 9.22 & 26.22 & 37.59 & 31.53 \\
\addlinespace[2pt]
\rowcolor{SwiftRow}
\textbf{SWIFT} & 1.50 & \textbf{54.36} & \textbf{9.70} & \underline{26.83} & \textbf{39.35} & \textbf{32.56} \\
\bottomrule
\end{tabularx}
\endgroup
\end{table}

\subsubsection{Image Classification}

To evaluate applicability beyond language tasks, we test SWIFT on five image classification datasets using CLIP ViT-B/32, as shown in Table~\ref{table3}. SWIFT achieves the highest average accuracy of $86.78\%$, outperforming full fine-tuning by 2.19 percentage points at a reported trainable-parameter ratio of $1.70\%$. It also surpasses LoRA by 8.09 percentage points, SpaRTA by 3.58 percentage points, and DiaBlo by 1.71 percentage points. SWIFT achieves the best accuracy on all five evaluated datasets, supporting the effectiveness of its update structure for visual adaptation.

\begin{table}[htbp]
\centering
\caption{Performance comparison of CLIP ViT-B/32 on five image classification datasets. Task scores are accuracies (\%). Bold and underlined entries indicate the best and second-best scores, respectively.}
\label{table3}
\begingroup
\fontsize{9}{11}\selectfont
\renewcommand{\arraystretch}{1.08}
\setlength{\tabcolsep}{3pt}
\begin{tabularx}{\linewidth}{@{}l*{7}{>{\centering\arraybackslash}X}@{}}
\toprule
\textbf{Method} & \textbf{\shortstack{Param.\\(\%)}} & \textbf{Cars} & \textbf{DTD} & \textbf{GTSRB} & \textbf{RESISC45} & \textbf{SVHN} & \textbf{Avg.} \\
\midrule
Full Fine-Tuning & 100 & 60.33 & 73.88 & \underline{98.30} & 93.65 & \underline{96.78} & 84.59 \\
\swiftgroup{8}{Low-rank adaptation}
LoRA & 1.49 & 41.02 & 70.15 & 96.51 & 90.38 & 95.39 & 78.69 \\
DoRA & 1.49 & 40.75 & 71.91 & 97.71 & 90.19 & 95.46 & 79.20 \\
PiSSA & 1.49 & 40.41 & 69.62 & 95.84 & 90.58 & 95.84 & 78.46 \\
GOAT & 2.24 & 51.44 & 73.99 & 98.06 & 93.50 & 96.55 & 82.71 \\
\swiftgroup{8}{Sparse and structured adaptation}
MoRe & 1.68 & 62.63 & \underline{74.26} & 98.16 & \underline{93.68} & 96.24 & 84.99 \\
SpaRTA & 1.60 & 58.40 & 70.40 & 97.90 & 92.90 & 96.40 & 83.20 \\
DiaBlo & 1.70 & \underline{63.67} & 73.67 & 97.98 & 93.56 & 96.49 & \underline{85.07} \\
\addlinespace[2pt]
\rowcolor{SwiftRow}
\textbf{SWIFT} & 1.70 & \textbf{67.69} & \textbf{76.06} & \textbf{98.57} & \textbf{94.46} & \textbf{97.10} & \textbf{86.78} \\
\bottomrule
\end{tabularx}
\endgroup
\end{table}

\subsection{Ablation Study}

\begin{figure}[htbp]
\centering
\includegraphics[width=\linewidth]{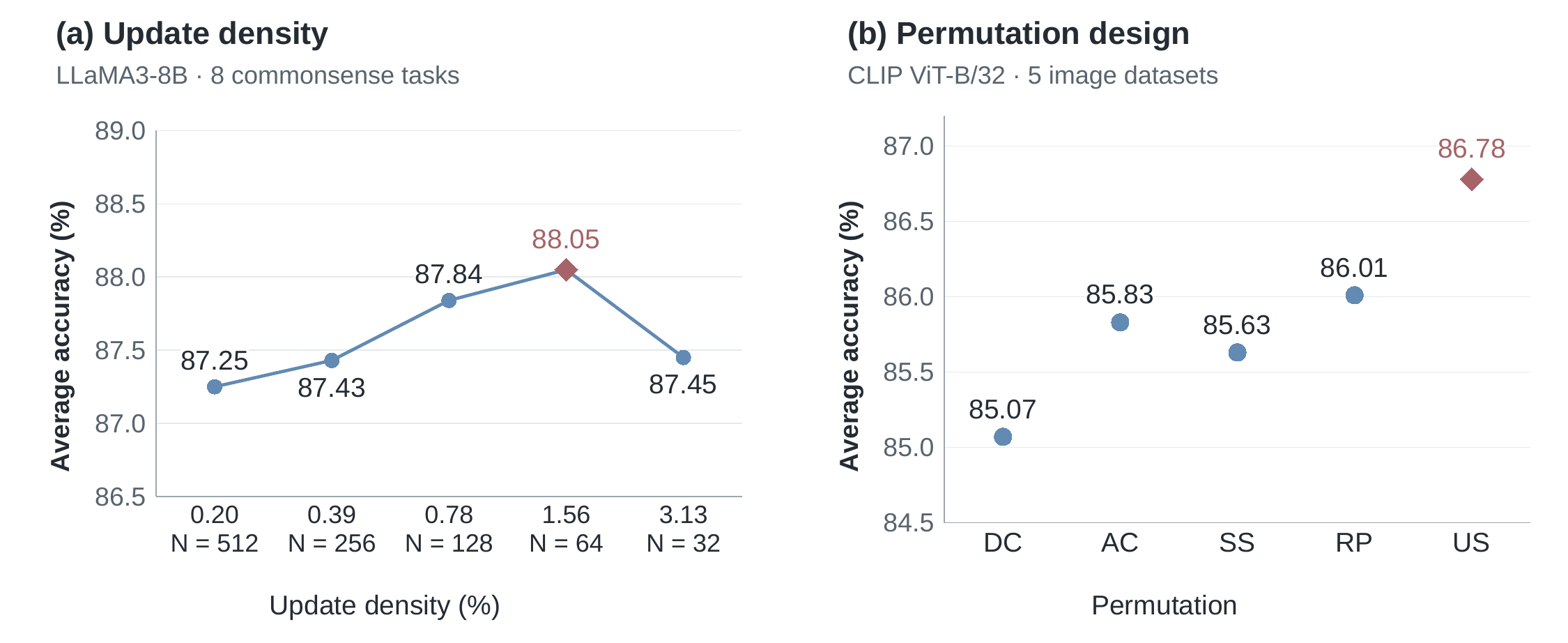}
\caption{Ablation study of SWIFT.
(a) Update-density sensitivity on eight commonsense reasoning datasets with LLaMA3-8B. Densities of $0.20\%$, $0.39\%$, $0.78\%$, $1.56\%$, and $3.13\%$ correspond to $N=512,256,128,64,32$, respectively.
(b) Comparison of permutation routing schemes on five image classification datasets with CLIP ViT-B/32. DC, AC, SS, RP, and US denote diagonal cyclic, anti-diagonal cyclic, single-row swap, random permutation, and uniform scattering, respectively.}
\label{Figure3}
\end{figure}

\paragraph{Update-density sensitivity.}
We study the sensitivity of SWIFT to the update density by varying the number of blocks $N$ in the block-diagonal update. When the input and output dimensions are divisible by $N$, the structural density of each adapted matrix is $1/N$. As shown in Figure~\ref{Figure3}(a), densities of $0.20\%$, $0.39\%$, $0.78\%$, $1.56\%$, and $3.13\%$ correspond to $N=512,256,128,64,32$, respectively. SWIFT achieves its highest average accuracy of $88.05\%$ at $N=64$. Increasing the update density beyond this setting does not further improve accuracy in this experiment. These results support $N=64$ as a practical default for the evaluated settings.

\paragraph{Permutation type.}
We further compare routing schemes in Figure~\ref{Figure3}(b), including diagonal cyclic (DC), anti-diagonal cyclic (AC), single-row swap (SS), random permutation (RP), and uniform scattering (US). US achieves the highest average accuracy of $86.78\%$, outperforming RP by 0.77 percentage points and DC by 1.71 percentage points. These results support the benefit of the deterministic interleaving construction, which distributes locally mixed outputs across feature groups.

\paragraph{Additional analyses.}
On LLaMA3-8B, SWIFT improves over DiaBlo by 1.24, 0.93, and 1.05 percentage points for attention-only, feed-forward-only, and combined placement, respectively, with matched budgets within each placement. Across seeds 6, 24, and 42, SWIFT achieves $88.05\%$ mean accuracy with a sample standard deviation of 0.10 percentage points. Appendix~\ref{app:additional_experiments} provides the complete placement and seed results, together with the training-loss comparison.

\subsection{Practical Efficiency}
\label{sec:practical_efficiency}

We conduct an additional efficiency comparison on LLaMA3-8B. All compared methods use a matched trainable-parameter budget of $0.88\%$ and are trained on a single NVIDIA Tesla V100 GPU with 32 GB of memory using FP16 precision. The comparison uses the same data, splits, and optimizer settings, with method-specific training durations: SWIFT is trained for two epochs and DiaBlo for three epochs. Results are averaged over seeds 6, 24, and 42. Memory cost denotes the maximum single-GPU training memory under the shared measurement protocol. Training time and FLOPs denote the total wall-clock time and cumulative floating-point operations over each method's complete training run.

\begin{figure}[htbp]
\centering
\includegraphics[width=\linewidth]{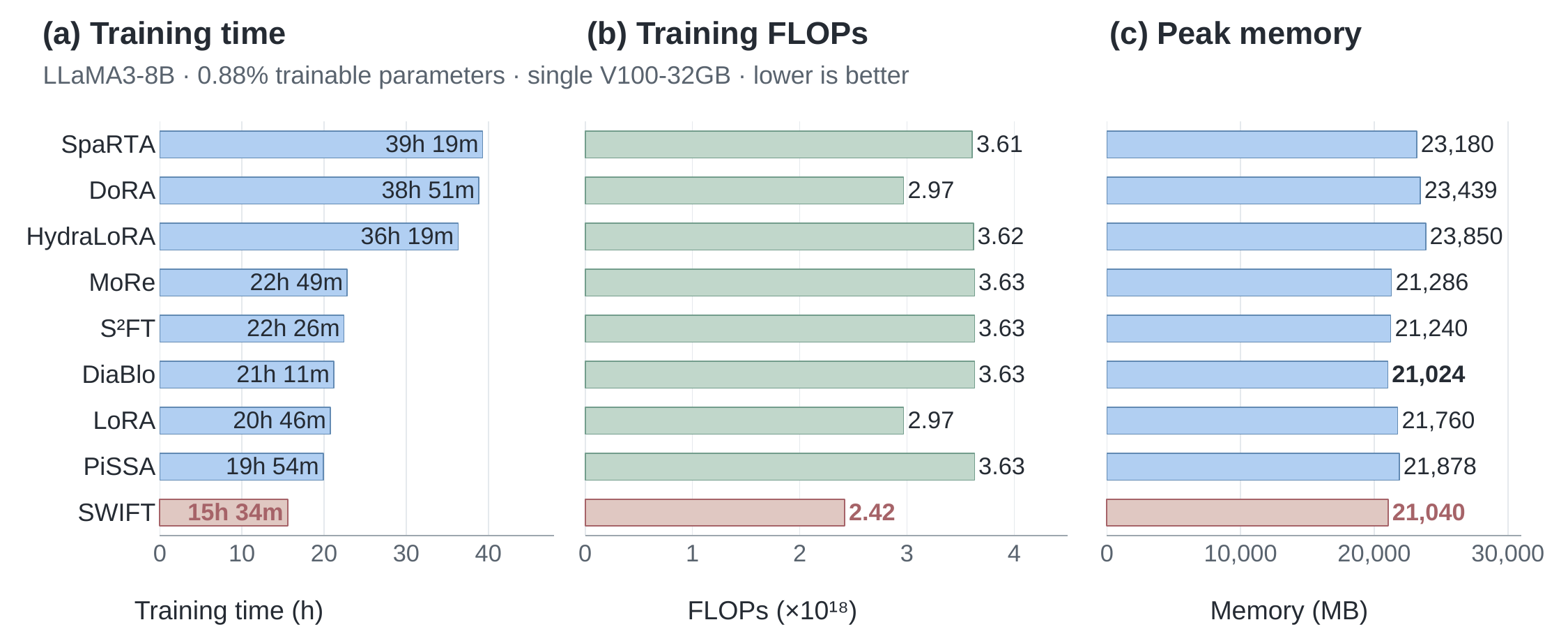}
\caption{End-to-end efficiency comparison on LLaMA3-8B at a matched $0.88\%$ trainable-parameter budget on one V100-32GB GPU. (a) Total training time. (b) Total training FLOPs. (c) Peak training memory. Training durations follow the method-specific schedules described in the text. Results are averaged over three seeds; lower values are better.}
\label{fig:efficiency}
\end{figure}

As shown in Figure~\ref{fig:efficiency}, under the respective training schedules, SWIFT has the shortest total training time and the lowest cumulative FLOP count among the compared implementations, reducing training time by $21.8$--$60.4\%$ and total training FLOPs by $18.5$--$33.3\%$ relative to the eight baselines. Its memory cost is 21,040 MB, only 16 MB above the minimum achieved by DiaBlo.

For merged-model inference, the effective weight is formed once as
$\mathbf{W}_{\mathrm{merged}}=\mathbf{W}+\mathbf{B}\mathbf{P}$,
after which the original dense linear layer is used. In the additional single-GPU LLaMA3-8B commonsense-reasoning inference comparison with batch size 1, SWIFT and the evaluated mergeable baselines achieve the same reported throughput of 6.83 samples/s.

\section{Conclusion}
We proposed SWIFT, a sparse fine-tuning framework inspired by small-world connectivity. By combining block-diagonal sparse updates with a fixed deterministic scattering permutation, SWIFT performs local feature mixing followed by cross-group routing under a low parameter budget. We further provided a topology-driven analysis showing that certain random sparse-update models and block-diagonal updates can exhibit information islands, whereas SWIFT guarantees positive block-level expansion under the stated structural and scattering conditions. Extensive experiments on commonsense reasoning, natural language generation, and image classification demonstrate that SWIFT achieves higher average performance than the evaluated low-rank and sparse fine-tuning baselines in each task category. Overall, SWIFT offers both an effective PEFT method and a graph-theoretic perspective for designing sparse update structures with improved cross-group connectivity.

\section{Limitations}
\label{sec:limitations}
SWIFT achieves strong results across language and vision tasks through structured scattering. Its theoretical expansion guarantees require equal-sized, structurally dense square blocks and a sufficient scattering condition. Building on this foundation, exploring more flexible connectivity patterns while preserving parameter efficiency, and extending the theoretical guarantees to more general settings, are promising directions for future work.

\label{page:main_end}
\clearpage

\section*{Acknowledgments}
This work was supported in part by the National Natural Science Foundation of China under Grant 62372132 and in part by the Major Key Project of Pengcheng Laboratory under Grant PCL2025A14. The authors would also like to thank Huawei Ascend Cloud Ecological Development Project for providing high-performance Ascend 910 processors. The authors declare no competing interests.

\bibliographystyle{unsrtnat}
\bibliography{reference}

\clearpage
\appendix

\section{Datasets and Experimental Settings}

\subsection{Datasets}
\label{A1}

\textbf{Commonsense Reasoning Tasks.}
We evaluate the commonsense reasoning capabilities of SWIFT on the following datasets:

\begin{enumerate}
    \item \textbf{BoolQ} \cite{clark2019boolq}: A question-answering dataset comprising yes/no questions.
    \item \textbf{PIQA} (Physical Interaction Question Answering) \cite{bisk2020piqa}: A dataset of questions accompanied by two candidate solutions, where identifying the correct answer requires physical commonsense reasoning.
    \item \textbf{SIQA} (Social IQA) \cite{sap2019socialiqa}: A question-answering dataset focused on reasoning about motivations, reactions, and social implications in everyday situations.
    \item \textbf{HellaSwag} \cite{zellers2019hellaswag}: A commonsense sentence-completion benchmark in which each context is accompanied by multiple candidate endings.
    \item \textbf{WinoGrande} \cite{sakaguchi2021winogrande}: A fill-in-the-blank benchmark with two candidate options for each sentence.
    \item \textbf{ARC-e/ARC-c} (AI2 Reasoning Challenge--Easy/Challenge) \cite{clark2018think}: The Easy and Challenge subsets of ARC, containing multiple-choice questions about school-level science.
    \item \textbf{OBQA} (OpenBookQA) \cite{mihaylov2018can}: A question-answering dataset requiring reasoning that combines scientific facts with additional common knowledge.
\end{enumerate}

Following \cite{hu2023llm}, we use Commonsense-170K for training. This dataset contains approximately 170,000 examples in a standardized instruction format, covering diverse aspects of commonsense reasoning.

\textbf{Natural Language Generation Tasks.}
We evaluate the natural language generation capabilities of SWIFT on the following datasets:

\begin{enumerate}
    \item \textbf{GSM8K} \cite{cobbe2021training}: A mathematical reasoning dataset comprising high-quality, linguistically diverse grade-school mathematical word problems written by human problem creators.
    \item \textbf{MATH} \cite{hendrycks2021math}: A mathematical reasoning dataset comprising competition mathematics problems across seven subject categories.
    \item \textbf{HumanEval} \cite{chen2021evaluating}: A code generation benchmark that evaluates the functional correctness of generated Python programs using unit tests.
    \item \textbf{MBPP} \cite{austin2021program}: A benchmark of crowdsourced Python programming problems intended to be approachable for novice programmers, covering basic programming concepts and standard library functionality.
\end{enumerate}

Following \cite{meng2024pissa}, we use the following two datasets for training:

\begin{enumerate}
    \item \textbf{Math: MetaMathQA} \cite{yu2023metamath}. MetaMathQA is a synthetic dataset developed to improve mathematical reasoning in large language models. We use 100K examples from the MetaMathQA training data prepared by PiSSA and evaluate the resulting model on GSM8K and MATH.
    \item \textbf{Code: Code-Feedback} \cite{zheng2024opencodeinterpreter}. We use 100K examples from the Python instruction-response data prepared by PiSSA from CodeFeedback and evaluate the resulting model on HumanEval and MBPP.
\end{enumerate}

For both training datasets, the released scripts select the first 100K examples from the corresponding training split.

\textbf{Image Classification Tasks.}
We evaluate SWIFT on the following image classification datasets:

\begin{enumerate}
    \item \textbf{Cars (Stanford Cars)} \cite{krause20133d}: A car classification dataset containing 16,185 images across 196 classes, with separate training and test sets.
    \item \textbf{DTD (Describable Textures Dataset)} \cite{cimpoi2014describing}: A texture classification dataset containing 5,640 images across 47 classes, with 120 images per class.
    \item \textbf{GTSRB (German Traffic Sign Recognition Benchmark)} \cite{stallkamp2011german}: A traffic sign classification dataset containing over 50,000 images across 43 categories.
    \item \textbf{RESISC45} \cite{cheng2017remote}: A remote sensing scene classification dataset containing 31,500 images across 45 categories, with 700 images per category.
    \item \textbf{SVHN (Street View House Numbers)} \cite{netzer2011reading}: A digit classification dataset derived from Google Street View images, containing 10 classes, with 73,257 training images, 26,032 test images, and an additional set of 531,131 images. Our implementation uses the official training and test splits without the additional set.
\end{enumerate}

Following \cite{fan2025make}, we report classification accuracy on all five datasets.

\subsection{Experimental Settings}
\label{A2}

\paragraph{Implementation Details.}
SWIFT is implemented in PyTorch using Hugging Face Transformers. We use LLaMA3-8B for commonsense reasoning, LLaMA2-7B for natural language generation, and CLIP ViT-B/32 for image classification. The released SWIFT training scripts default to one GPU with FP16 precision for commonsense reasoning, two GPUs with BF16 precision for natural language generation, and four GPUs with BF16 mixed precision for image classification. The distributed generation and image-classification configurations use DeepSpeed ZeRO Stage 2.

The pre-trained backbone weights remain frozen during SWIFT fine-tuning. The trainable block-diagonal updates are initialized to zero, and the scattering permutation is fixed throughout training. For image classification, the classification head is trained jointly with the SWIFT updates. 

All downstream experiments, including the main comparisons, ablations, and efficiency evaluations, use random seeds 6, 24, and 42, and the reported results are averaged across the three runs.

\paragraph{Hyperparameters.}
For SWIFT, the number of blocks is set to 64 in all main experiments; the block-number ablation varies this setting explicitly. We use the same deterministic scattering construction across tasks without task-specific permutation search. SWIFT uses AdamW with zero weight decay and does not apply dropout within its update branch. Detailed task-specific settings are provided below.

For commonsense reasoning, full fine-tuning uses a learning rate of $1\times10^{-5}$. LoRA, DoRA, PiSSA, and HydraLoRA use a learning rate of $3\times10^{-4}$ with rank 32. SpIEL and SMT follow the hyperparameter settings reported in the SMT paper, and S$^{2}$FT follows the settings reported in the S$^{2}$FT paper. SpaRTA, DiaBlo, MoRe, and SWIFT use a learning rate of $5\times10^{-5}$.

SWIFT adapts the query, key, and value projections and the feed-forward up and down projections, corresponding to \texttt{q\_proj}, \texttt{k\_proj}, \texttt{v\_proj}, \texttt{up\_proj}, and \texttt{down\_proj}. Training lasts two epochs with a global batch size of 16, a per-device micro-batch size of 2, and eight gradient accumulation steps in the default single-GPU configuration. The maximum sequence length is 256 tokens, and the learning-rate schedule uses 100 warmup steps. Prompt tokens are excluded from the training loss. We hold out 120 examples from the training data for validation using split seed 42. Validation and checkpoint saving occur every 80 optimizer steps, and the checkpoint with the lowest validation loss is selected.

For natural language generation, full fine-tuning follows the hyperparameter settings used in the PiSSA paper. LoRA, DoRA, and PiSSA use a learning rate of $2\times10^{-5}$ with rank 160, following PiSSA. HydraLoRA uses a learning rate of $2\times10^{-5}$ with rank 32. SpaRTA, DiaBlo, MoRe, and SWIFT use a learning rate of $2\times10^{-4}$.

SWIFT adapts all attention projections and the feed-forward gate, up, and down projections, corresponding to \texttt{q\_proj}, \texttt{k\_proj}, \texttt{v\_proj}, \texttt{o\_proj}, \texttt{gate\_proj}, \texttt{up\_proj}, and \texttt{down\_proj}. Math and code models are fine-tuned separately for one epoch. The maximum sequence length is 512 tokens. The default configuration uses a per-device micro-batch size of 1 and two gradient accumulation steps on two GPUs, giving a global batch size of 4. We use a cosine learning-rate schedule with a warmup ratio of 0.03 and evaluate the final checkpoint.

For image classification, full fine-tuning, LoRA, DoRA, and PiSSA follow the hyperparameter settings used in the GOAT paper, with a learning rate of $1\times10^{-4}$ and rank 8 for the low-rank methods. SpaRTA, DiaBlo, and MoRe use a learning rate of $1\times10^{-3}$. SWIFT uses a learning rate of $1\times10^{-4}$.

SWIFT adapts the linear layers of the vision backbone, including the attention and feed-forward projections. The classification head is trained directly. Training lasts 35, 76, 11, 15, and 4 epochs for Cars, DTD, GTSRB, RESISC45, and SVHN, respectively. The default configuration uses a global batch size of 512, a per-device micro-batch size of 64, and two gradient accumulation steps on four GPUs. We use a cosine learning-rate schedule with a warmup ratio of 0.03.

The vision data loader reserves 10\% of each supplied training split, capped at 5,000 examples, using split seed 0, and uses the remaining examples for optimization. Training images undergo random resized cropping to $224\times224$ pixels, with a crop scale of $[0.9,1.0]$, followed by CLIP normalization. Evaluation images are resized and center-cropped to $224\times224$ pixels using the same normalization. Evaluation accuracy is recorded at the end of each epoch, and the final adapter and classification head are saved after training. All SWIFT training scripts complete the specified epoch schedule without early stopping.

\paragraph{Evaluation Settings.}
For commonsense reasoning, the released evaluation script uses a batch size of 1, four-beam decoding, and a maximum of 32 generated tokens. Predictions are extracted using the task-specific answer formats and evaluated by accuracy. For natural language generation, the math and code evaluation scripts use greedy decoding with a batch size of 4 and a maximum of 1,024 generated tokens. Mathematical answers are extracted and compared with the references using the provided evaluation utilities. Code generation is evaluated with EvalPlus using one generated completion per problem.

\section{Proof of Small-World Induced Fine-Tuning Theory}

This appendix proves Theorems~\ref{thm:nonregular}, \ref{thm:block}, and \ref{thm:swift}, together with the MRGC bound in Equation~\eqref{eq:mrgc_scattering} and the coordinate-level expansion bound in Equation~\eqref{eq:coordinate_expansion}. Throughout, graphs are induced by the structural support of trainable updates, including trainable entries initialized to zero. Under the row-vector convention, an edge $(i,j)$ represents a possible direct trainable interaction from input coordinate $i$ to output coordinate $j$. We use
\[
O(S)=\{j\in V:\exists i\in S,\ (i,j)\in E\},
\qquad
\partial_{\mathrm{out}}S=O(S)\setminus S.
\]
The quantities $h(G)$ and $h_{\mathcal{P}}(G)$, defined in Equations~\eqref{eq:cheeger} and \eqref{eq:block_expansion}, measure directed out-vertex expansion over coordinate subsets and block unions, respectively.

\subsection{Additional Coverage and Expansion Bounds}
\label{app:additional_bounds}

The inverse permutation satisfies the same scattering bound, since
\[
|\pi^{-1}(V_r)\cap V_q|
=
|\pi(V_q)\cap V_r|
\le \epsilon m.
\]
Consequently, this condition controls the output routing under the convention in Equation~\ref{eq:permutation_convention}.

For a union $U_I$ of $s$ blocks, structural density gives $O(U_I)=\pi^{-1}(U_I)$ and $|O(U_I)|=sm$. The scattering condition bounds its overlap with $U_I$ by $s^2\epsilon m$, yielding Equation~\ref{eq:swift_expansion}. The same condition also provides a separate remote-group coverage bound:
\begin{equation}
\mathrm{MRGC}
\ge
\frac{\lceil 1/\epsilon\rceil-1}{k-1}.
\label{eq:mrgc_scattering}
\end{equation}
Indeed, every source coordinate has $m$ output neighbors, with at most $\epsilon m$ in any target block, so it reaches at least $\lceil 1/\epsilon\rceil$ blocks, at most one of which is local.

\paragraph{Coordinate-level expansion.}
Under the assumptions of Theorem~\ref{thm:swift}, a separate coordinate-level bound holds:
\begin{equation}
h(G_{\mathbf{B}\mathbf{P}})
\ge
\min\left\{
1-\frac{k\epsilon}{2},\ c(k)
\right\}
>0,
\qquad
c(k)=
\begin{cases}
2/k, & k\text{ even},\\
1/k, & k\text{ odd}.
\end{cases}
\label{eq:coordinate_expansion}
\end{equation}
This follows by considering how many blocks a coordinate subset intersects and using the structurally dense support within each block. The block-level theorem, MRGC bound, and coordinate-level bound are derived separately from the same scattering condition. Detailed proofs are provided in Appendix~\ref{B3}.

Under exact uniform scattering, $\epsilon=1/k$, Theorem~\ref{thm:swift} gives $h_{\mathcal{P}}(G_{\mathbf{B}\mathbf{P}})\ge 1/2$. For other compatible dimensions, the realized value in Equation~\ref{eq:real_scattering} determines whether the sufficient condition $\epsilon_{\mathrm{real}}<2/k$ holds. Permutations that only exchange entire blocks have $\epsilon=1$; in particular, swapping disjoint block pairs leaves closed block unions when $k\ge4$. The expansion guarantees above apply to equal-sized, structurally dense square blocks, while rectangular projections are characterized through their bipartite group coverage.

\subsection{Proof of Theorem~\ref{thm:nonregular}: Connectivity Collapse under Independent Random Sparsification}
\label{B1}

Let $G_d$ be the interaction graph induced by the random structural support, and let $D_{\mathrm{out}}(v)$ denote the out-degree of vertex $v$. Write
\[
p_{v,d}=\mathbb{P}\bigl(D_{\mathrm{out}}(v)=0\bigr).
\]
The uniform lower-bound assumption
\[
\liminf_{d\to\infty}
\min_{v\in\{1,\ldots,d\}}p_{v,d}>0
\]
implies that there exist constants $\eta\in(0,1)$ and $d_0$ such that
\begin{equation}
p_{v,d}\ge\eta
\qquad
\text{for every }d\ge d_0
\text{ and every }v\in\{1,\ldots,d\}.
\label{eq:app_uniform_zero_degree}
\end{equation}

Let
\[
X_d=\sum_{v=1}^{d}
\mathbb{I}_{\{D_{\mathrm{out}}(v)=0\}}
\]
be the number of zero-out-degree vertices. Since outgoing supports are sampled independently across vertices, the events $\{D_{\mathrm{out}}(v)=0\}$ are independent. Therefore, for $d\ge d_0$,
\begin{equation}
\mathbb{P}(X_d=0)
=
\prod_{v=1}^{d}(1-p_{v,d})
\le
(1-\eta)^d
\xrightarrow[d\to\infty]{}0.
\label{eq:app_no_zero_degree}
\end{equation}
Thus, $\mathbb{P}(X_d\ge1)\to1$.

For $d\ge2$, if a vertex $v$ has zero out-degree, then the singleton $S=\{v\}$ is admissible in Equation~\eqref{eq:cheeger} and satisfies
\[
\partial_{\mathrm{out}}S=\emptyset.
\]
Consequently,
\[
0\le h(G_d)
\le
\frac{|\partial_{\mathrm{out}}S|}{|S|}
=0.
\]
It follows that
\begin{equation}
\mathbb{P}\bigl(h(G_d)=0\bigr)
\ge
\mathbb{P}(X_d\ge1)
\ge
1-(1-\eta)^d
\xrightarrow[d\to\infty]{}1.
\end{equation}
This proves Theorem~\ref{thm:nonregular}. \hfill $\blacksquare$

For the independent Bernoulli model with entry probability $a/d$, the expected out-degree is $a$ and
\begin{equation}
p_{v,d}
=
\left(1-\frac{a}{d}\right)^d
\xrightarrow[d\to\infty]{}
e^{-a}>0.
\end{equation}
Hence this model satisfies the theorem's assumptions. The proof applies more generally whenever outgoing supports are independent across vertices and the zero-out-degree probabilities satisfy the stated uniform lower bound.

\subsection{Proof of Theorem~\ref{thm:block}: Zero Expansion of Block-Diagonal Updates}
\label{B2}

Let $\mathbf{B}\in\mathbb{R}^{d\times d}$ have block-diagonal structural support under the partition
\[
V=\bigsqcup_{r=1}^{k}V_r,
\qquad
|V_r|=m,
\qquad
d=km,
\qquad
k>1.
\]
Thus,
\[
\Omega_{\mathbf{B}}
\subseteq
\bigcup_{r=1}^{k}(V_r\times V_r).
\]
Consider $S=V_1$. Since $k>1$,
\[
1\le |S|=m\le\lfloor d/2\rfloor.
\]
For any source vertex $i\in S$, an edge $(i,j)\in\Omega_{\mathbf{B}}$ requires $j$ to belong to the same block as $i$. Therefore $O(S)\subseteq S$, giving
\[
\partial_{\mathrm{out}}S
=
O(S)\setminus S
=
\emptyset.
\]
Substituting this admissible set into Equation~\eqref{eq:cheeger} yields
\[
0
\le
h(G_{\mathbf{B}})
\le
\frac{|\partial_{\mathrm{out}}S|}{|S|}
=
0.
\]
Hence $h(G_{\mathbf{B}})=0$, proving Theorem~\ref{thm:block}. \hfill $\blacksquare$

\subsection{Proof of Theorem~\ref{thm:swift}: Expansion of Small-World Induced Fine-Tuning}
\label{B3}

Let $\mathcal{P}=\{V_r\}_{r=1}^{k}$ be a partition of $V$ into $k>1$ blocks of equal size $m=d/k$. The trainable block-diagonal factor has structurally dense blocks:
\[
\Omega_{\mathbf{B}}
=
\bigcup_{r=1}^{k}(V_r\times V_r).
\]
Let $\mathbf{P}$ be the permutation matrix associated with $\pi$ under the convention
\begin{equation}
(\mathbf{B}\mathbf{P})_{ij}
=
\mathbf{B}_{i,\pi(j)},
\label{eq:bp_reindex_proof}
\end{equation}
and write
\[
\Delta\mathbf{W}=\mathbf{B}\mathbf{P},
\qquad
G=G_{\mathbf{B}\mathbf{P}}.
\]

\paragraph{Output neighborhoods and inverse scattering.}
For any source coordinate $i\in V_r$,
\[
(i,j)\in\Omega_{\Delta}
\quad\Longleftrightarrow\quad
(i,\pi(j))\in\Omega_{\mathbf{B}}
\quad\Longleftrightarrow\quad
\pi(j)\in V_r.
\]
Structural density therefore gives the exact neighborhood identity
\begin{equation}
O(\{i\})=\pi^{-1}(V_r).
\label{eq:app_single_neighborhood}
\end{equation}
In particular, every source coordinate has exactly $m$ neighbors in the structural support.

The scattering assumption also holds for $\pi^{-1}$. Indeed, for all $r,q\in\{1,\ldots,k\}$,
\begin{equation}
|\pi^{-1}(V_r)\cap V_q|
=
|V_r\cap\pi(V_q)|
\le
\epsilon m.
\label{eq:app_inverse_scattering}
\end{equation}

\paragraph{Block-level expansion.}
Consider any admissible block union
\[
U_I=\bigcup_{r\in I}V_r,
\qquad
1\le |I|=s\le\lfloor k/2\rfloor.
\]
By Equation~\eqref{eq:app_single_neighborhood},
\begin{equation}
O(U_I)
=
\bigcup_{r\in I}\pi^{-1}(V_r)
=
\pi^{-1}(U_I).
\label{eq:app_union_neighborhood}
\end{equation}
Since $\pi$ is a bijection,
\begin{equation}
|O(U_I)|=|U_I|=sm.
\label{eq:app_union_size}
\end{equation}

The sets $\{\pi^{-1}(V_r)\}_{r=1}^{k}$ form a partition of $V$, as do the sets $\{V_q\}_{q=1}^{k}$. Consequently, the following intersections are pairwise disjoint, and Equation~\eqref{eq:app_inverse_scattering} gives
\begin{equation}
\begin{aligned}
|O(U_I)\cap U_I|
&=
|\pi^{-1}(U_I)\cap U_I|\\
&=
\sum_{r\in I}\sum_{q\in I}
|\pi^{-1}(V_r)\cap V_q|\\
&\le
s^2\epsilon m.
\end{aligned}
\label{eq:overlap_bound}
\end{equation}
Combining Equations~\eqref{eq:app_union_size} and \eqref{eq:overlap_bound},
\begin{equation}
\begin{aligned}
|\partial_{\mathrm{out}}U_I|
&=
|O(U_I)|-|O(U_I)\cap U_I|\\
&\ge
sm-s^2\epsilon m\\
&=
sm(1-s\epsilon).
\end{aligned}
\end{equation}
Dividing by $|U_I|=sm$ yields
\begin{equation}
\frac{|\partial_{\mathrm{out}}U_I|}{|U_I|}
\ge
1-s\epsilon
\ge
1-\frac{k\epsilon}{2}.
\end{equation}
Taking the minimum over all admissible block unions gives
\begin{equation}
h_{\mathcal{P}}(G)
\ge
1-\frac{k\epsilon}{2}
=
1-\frac{d\epsilon}{2m}.
\end{equation}
Under the sufficient condition $\epsilon<2/k$, this lower bound is strictly positive. This proves Theorem~\ref{thm:swift}. \hfill $\blacksquare$

\paragraph{Proof of the MRGC bound.}
We next prove Equation~\eqref{eq:mrgc_scattering} using the same partition, so that $K=k$ in the MRGC definition. Fix a source coordinate $i\in V_r$, and let
\[
\mathcal{L}(i)
=
\{q\in\{1,\ldots,k\}:O(\{i\})\cap V_q\neq\emptyset\}
\]
be the set of all target groups reached by $i$, including its local group if reached. By Equations~\eqref{eq:app_single_neighborhood} and \eqref{eq:app_inverse_scattering},
\[
|O(\{i\})|=m,
\qquad
|O(\{i\})\cap V_q|\le\epsilon m
\quad\text{for every }q.
\]
Therefore,
\[
m
=
\sum_{q\in\mathcal{L}(i)}
|O(\{i\})\cap V_q|
\le
|\mathcal{L}(i)|\,\epsilon m,
\]
which implies
\[
|\mathcal{L}(i)|\ge\lceil1/\epsilon\rceil.
\]
Excluding the local group removes at most one group, so
\[
|\mathcal{R}(i)|
\ge
\lceil1/\epsilon\rceil-1.
\]
Since this holds for every source coordinate,
\begin{equation}
\mathrm{MRGC}
=
\min_{i\in V}
\frac{|\mathcal{R}(i)|}{k-1}
\ge
\frac{\lceil1/\epsilon\rceil-1}{k-1}.
\end{equation}
This proves Equation~\eqref{eq:mrgc_scattering}. \hfill $\blacksquare$

\paragraph{Proof of the coordinate-level expansion bound.}
We finally prove Equation~\eqref{eq:coordinate_expansion}. Let $S\subseteq V$ be any set satisfying
\[
1\le |S|\le\lfloor d/2\rfloor.
\]
Define
\[
I_S=\{r:S\cap V_r\neq\emptyset\},
\qquad
t=|I_S|,
\qquad
U_{I_S}=\bigcup_{r\in I_S}V_r.
\]
Every coordinate in a given source block has the same output neighborhood by Equation~\eqref{eq:app_single_neighborhood}. Since $S$ intersects every block indexed by $I_S$,
\begin{equation}
O(S)
=
O(U_{I_S})
=
\pi^{-1}(U_{I_S}),
\qquad
|O(S)|=tm.
\label{eq:app_arbitrary_neighborhood}
\end{equation}
We consider two cases.

\emph{Case 1: $t\le k/2$.}
Since $S\subseteq U_{I_S}$, the overlap estimate in Equation~\eqref{eq:overlap_bound}, applied to $U_{I_S}$, gives
\[
|O(S)\cap S|
\le
|\pi^{-1}(U_{I_S})\cap U_{I_S}|
\le
t^2\epsilon m.
\]
Consequently,
\[
|\partial_{\mathrm{out}}S|
=
|O(S)|-|O(S)\cap S|
\ge
tm(1-t\epsilon).
\]
The assumption $\epsilon<2/k$ implies
\[
1-t\epsilon
\ge
1-\frac{k\epsilon}{2}
>0.
\]
Using $|S|\le tm$, we obtain
\begin{equation}
\frac{|\partial_{\mathrm{out}}S|}{|S|}
\ge
\frac{tm(1-t\epsilon)}{|S|}
\ge
1-t\epsilon
\ge
1-\frac{k\epsilon}{2}.
\label{eq:app_coordinate_case_one}
\end{equation}

\emph{Case 2: $t>k/2$.}
Using Equation~\eqref{eq:app_arbitrary_neighborhood} and the bound $|O(S)\cap S|\le|S|$, we have
\[
|\partial_{\mathrm{out}}S|
\ge
tm-|S|.
\]
Since $|S|\le d/2=km/2$,
\begin{equation}
\frac{|\partial_{\mathrm{out}}S|}{|S|}
\ge
\frac{tm}{|S|}-1
\ge
\frac{2t}{k}-1.
\label{eq:app_coordinate_case_two}
\end{equation}
The integer $t$ satisfies $t\ge\lfloor k/2\rfloor+1$, and hence
\[
\frac{2t}{k}-1
\ge
c(k),
\qquad
c(k)=
\begin{cases}
2/k, & k\text{ even},\\
1/k, & k\text{ odd}.
\end{cases}
\]

Combining the two cases, every admissible coordinate subset $S$ satisfies
\[
\frac{|\partial_{\mathrm{out}}S|}{|S|}
\ge
\min\left\{
1-\frac{k\epsilon}{2},\ c(k)
\right\}.
\]
Taking the minimum over $S$ gives
\begin{equation}
h(G_{\mathbf{B}\mathbf{P}})
\ge
\min\left\{
1-\frac{k\epsilon}{2},\ c(k)
\right\}
>0.
\end{equation}
This proves Equation~\eqref{eq:coordinate_expansion}. \hfill $\blacksquare$

\section{Additional Analyses}
\label{app:additional_experiments}

\subsection{Baseline Descriptions}
\label{app:baselines}

To assess the effectiveness of SWIFT, we perform a comprehensive evaluation against several baseline methods, grouped into three categories:

\begin{enumerate}
    \item \textbf{Full Fine-Tuning:}
    This baseline updates all pre-trained parameters, providing a reference for adaptation without restricting the trainable parameter budget.

    \item \textbf{Low-Rank Adaptation Family:}
    This group includes LoRA \citep{hu2022lora} and its variants. We include DoRA \citep{liu2024dora} for weight-decomposed adaptation, PiSSA \citep{meng2024pissa} for principal-component initialization, GOAT \citep{fan2025make} for SVD-structured mixture-of-experts adaptation and optimization alignment, and HydraLoRA \citep{tian2024hydralora} for asymmetric adaptation to heterogeneous data. These methods achieve parameter efficiency through low-rank parameterizations and their extensions.

    \item \textbf{Sparse and Structured Fine-Tuning Approaches:}
    This group includes random, dynamic, and structured tuning methods. We compare SWIFT against SpaRTA \citep{aliaga2025sparsity}, which updates randomly selected parameters, and S$^{2}$FT \citep{yang2024s}, which selects attention heads and feed-forward channels for structured sparse updates. We also include SMT \citep{he2025smt}, which selects task-relevant submatrices using saliency signals, and SpIEL \citep{ansell2024scaling}, which dynamically reallocates sparse trainable parameters. Finally, we compare with MoRe \citep{tan2024more}, which uses Monarch-style structured parameterizations with fixed permutations, and DiaBlo \citep{gurses2025diablo}, which fine-tunes diagonal blocks of selected weight matrices.
\end{enumerate}

\subsection{Support Simulation and Rectangular Coverage}
\label{app:support_simulation}

For rectangular layers, we use separate input and output vertex sets with $N$ paired input and output groups. The remote target groups for an input coordinate are the output groups other than its paired group, and the normalization is $N-1$. The square-graph expansion results in Section~\ref{3.2} apply to square layers.

We use MRGC to compare representative structural supports under a matched sparse budget, with full fine-tuning as a dense reference. We simulate $d=4096$ feature dimensions partitioned into $K=64$ equal-sized groups. The random sparse baseline uses the same expected density as block-diagonal tuning, $p=1/64$, and results are averaged over $1{,}000$ sampled masks. Full fine-tuning uses a dense support, block-diagonal tuning retains only within-group interactions, and SWIFT uses the support of $\mathbf{B}\mathbf{P}$ with the deterministic interleaving permutation.

Full fine-tuning achieves the maximal MRGC of $1.00$, while block-diagonal tuning yields $0.00$ due to the absence of cross-group interactions. Random sparse tuning improves coverage but remains uneven, achieving a mean MRGC of $0.41$. In this setting, SWIFT attains an MRGC of $1.00$ at density $1/64$. Thus, the scattering permutation provides complete remote-group coverage in this configuration without increasing the trainable parameter budget.

\subsection{Placement and Seed Analyses}
\label{app:placement_seeds}

\begin{figure}[htbp]
\centering
\includegraphics[width=\linewidth]{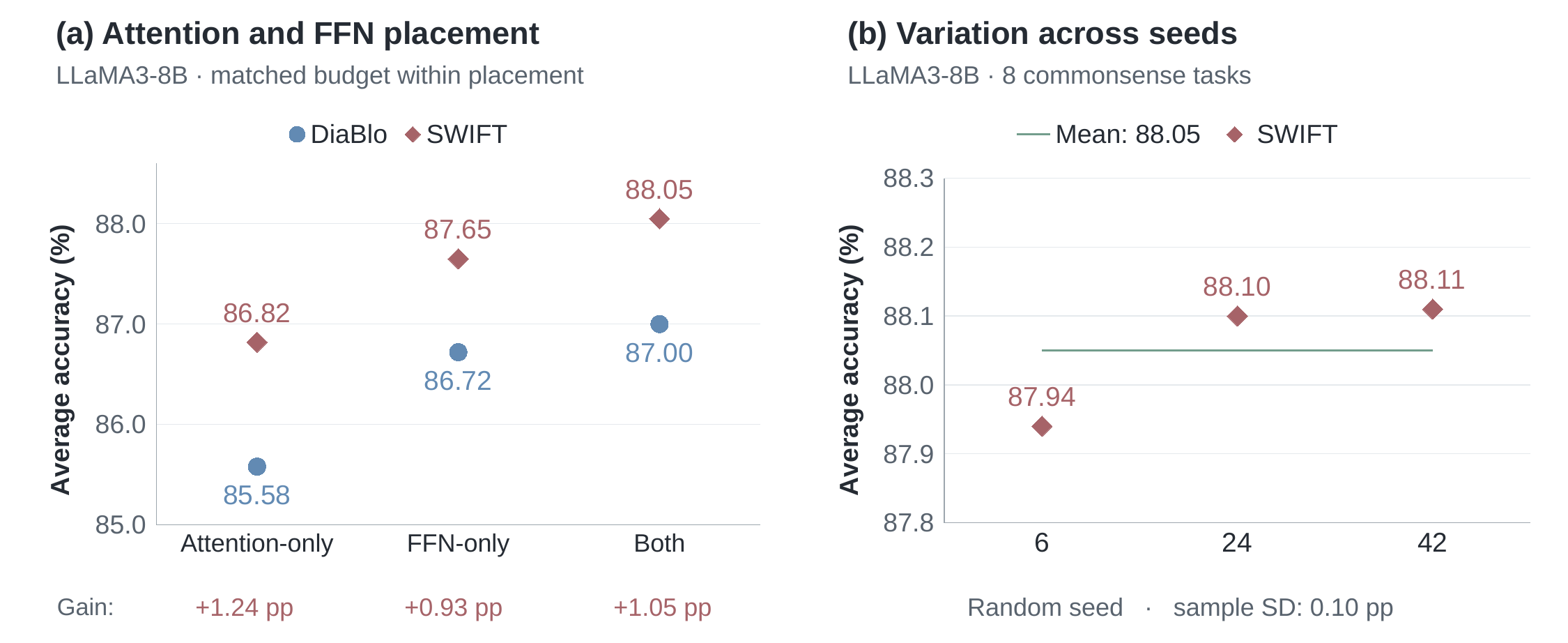}
\caption{Placement and seed analyses on LLaMA3-8B. (a) SWIFT and DiaBlo use identical target modules and matched parameter budgets within each placement; scores are averaged over eight commonsense reasoning datasets and three seeds. (b) SWIFT accuracy for seeds 6, 24, and 42, with the horizontal line showing their mean.}
\label{fig:placement_seeds}
\end{figure}

\paragraph{Attention and feed-forward placement.}
We additionally compare SWIFT and DiaBlo when updates are applied to attention projections, feed-forward projections, or both. Within each placement, the two methods use identical target modules and matched trainable-parameter budgets. Results are averaged over the eight commonsense reasoning datasets and seeds 6, 24, and 42. As shown in Figure~\ref{fig:placement_seeds}(a), SWIFT improves average accuracy by 1.24, 0.93, and 1.05 percentage points, respectively. The improvement across all three placements supports the applicability of permutation-based routing to both module families.

\paragraph{Variation across random seeds.}
\label{sec:seed_variation}

We additionally evaluate SWIFT on LLaMA3-8B using seeds 6, 24, and 42. The average accuracies across the eight commonsense reasoning datasets are $87.94\%$, $88.10\%$, and $88.11\%$, respectively, as shown in Figure~\ref{fig:placement_seeds}(b). The mean accuracy is $88.05\%$, with a sample standard deviation of approximately 0.10 percentage points, computed across the three seed-level averages using the $n-1$ denominator with $n=3$. These statistics describe variation across random seeds in this setting. No additional statistical significance test was performed.

\subsection{Convergence Analysis}

\begin{figure}[htbp]
\centering
\includegraphics[width=0.68\linewidth]{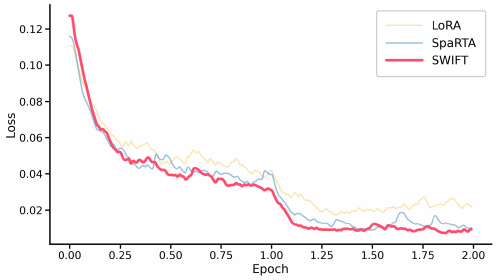}
\caption{Training loss curves for LoRA, SpaRTA, and SWIFT when fine-tuning LLaMA3-8B on Commonsense-170K for commonsense reasoning.}
\label{figure2}
\end{figure}

As shown in Figure~\ref{figure2}, we compare the training loss curves of SWIFT, LoRA, and SpaRTA over two epochs. SWIFT reaches a lower final training loss than both baselines and maintains a low loss during the later stage of training. These curves support favorable optimization behavior under the evaluated training schedule. Wall-clock training efficiency is assessed separately in Section~\ref{sec:practical_efficiency}.

\section{Broader Impacts and Asset Information}

\subsection{Broader Impacts}
\label{app:broader_impacts}

SWIFT reduces the number of trainable parameters required for adapting pre-trained models and demonstrates lower training costs in the evaluated efficiency comparison. These benefits may make task-specific adaptation more accessible to researchers with limited computational resources. Adapted models may inherit biases, factual inaccuracies, or other undesirable behaviors from their pre-trained backbones and training data. Parameter-efficient adaptation alone does not address these issues. Downstream deployment therefore requires application-specific evaluation and safeguards appropriate to the intended use.

\subsection{Asset Attribution and Licenses}
\label{app:asset_licenses}

The original model and dataset publications are cited in the main text and Appendix~\ref{A1}. LLaMA3-8B is distributed under the
\href{https://github.com/meta-llama/llama3/blob/main/LICENSE}{Meta Llama 3 Community License Agreement},
and LLaMA2-7B under the
\href{https://github.com/meta-llama/llama/blob/llama_v2/LICENSE}{Llama 2 Community License Agreement}.
The official
\href{https://github.com/openai/CLIP/blob/main/LICENSE}{CLIP implementation}
uses the MIT License, and
\href{https://github.com/huggingface/peft/blob/main/LICENSE}{PEFT}
uses the Apache License 2.0.

The SWIFT repository retains upstream license files and third-party notices, including the MIT license for the GOAT-derived image-classification code. Applicable subdirectory and file-level notices remain in effect. Pre-trained model weights and datasets are obtained separately from their original providers under the corresponding access conditions and terms of use; they are not redistributed with the SWIFT code.

\subsection{Code Availability}
\label{app:code_availability}
The SWIFT implementation is available at \url{https://github.com/xianchaoguan/SWIFT}, including training and evaluation scripts, environment specifications, and data-preparation instructions for commonsense reasoning, natural language generation, and image classification. The release covers the SWIFT pipelines for these three task groups; it does not include dedicated reproduction scripts for all baseline comparisons, ablations, and efficiency measurements. Pre-trained weights and datasets must be obtained separately from their original sources under the applicable access conditions.

\end{document}